\documentclass{article}
\usepackage{iclr2026_conference,times}

\usepackage{amsmath,amsfonts,bm}

\def\eqref#1{equation~\ref{#1}}

\def\1{\bm{1}}

\def\eps{{\epsilon}}

\DeclareMathAlphabet{\mathsfit}{\encodingdefault}{\sfdefault}{m}{sl}
\SetMathAlphabet{\mathsfit}{bold}{\encodingdefault}{\sfdefault}{bx}{n}

\newcommand{\E}{\mathbb{E}}

\newcommand{\R}{\mathbb{R}}

\newcommand{\KL}{D_{\mathrm{KL}}}

\usepackage[utf8]{inputenc}
\usepackage[T1]{fontenc}
\usepackage{lmodern}
\usepackage{microtype}
\usepackage{amsmath,amssymb,amsthm,mathtools,bm}
\usepackage{graphicx}
\usepackage{booktabs}
\usepackage{multirow}
\usepackage{enumitem}
\usepackage{hyperref}
\usepackage[capitalize,noabbrev]{cleveref}
\usepackage{xcolor}
\usepackage{array}
\hypersetup{colorlinks=true,linkcolor=blue!50!black,citecolor=blue!50!black,urlcolor=blue!50!black}
\newtheorem{theorem}{Theorem}
\newtheorem{lemma}{Lemma}
\newtheorem{proposition}{Proposition}

\newtheorem{assumption}{Assumption}
\newtheorem{definition}{Definition}
\providecommand{\R}{\mathbb{R}}
\providecommand{\E}{\mathbb{E}}
\providecommand{\Pbb}{\mathbb{P}}
\providecommand{\F}{\mathcal{F}}
\providecommand{\KL}{D_{\mathrm{KL}}}
\providecommand{\sgn}{\operatorname{sign}}

\providecommand{\Corr}{\operatorname{Corr}}
\providecommand{\HT}{\mathrm{HT}}
\providecommand{\eps}{\varepsilon}

\title{Dimension-Corrected Hitting Times for Heavy-Tailed Spectral Emergence in Neural Optimizer Dynamics}
\author{Zongmin Liu\\Stanford University\\\texttt{zongminl@stanford.edu}}
\iclrfinalcopy
\begin{document}
\maketitle
\begin{abstract}
Heavy-tailed empirical spectral densities (ESDs) of neural-network weight matrices are widely used as diagnostics of implicit self-regularization and representation learning.  Recent noise-free analyses show that heavy-tailed spectra can arise even without stochastic-gradient noise: large full-batch optimizer steps can first create an outlier spike beyond the Marchenko--Pastur bulk and later deform the bulk into a heavy-tailed shape.  What remains unexplained is the \emph{step complexity} of this spike-to-tail transition.  We formulate heavy-tail spectral emergence as a right-censored hitting-time problem.  Let \(\Delta_1\) be the first-step spike--bulk gap of \(d^{-1}W_1^\top W_1\), and let \(\tau_{\HT}\) be the first checkpoint at which a consensus heavy-tail diagnostic is satisfied.  Across 330 completed full-batch teacher--student runs, including the original 255-run design and robustness grids, finite-onset regression supports a dimension-corrected law
\[
\tau_{\HT}\approx C\Delta_1^{-\gamma}d^\rho,
\qquad R^2=0.683,\quad \gamma=0.626,\quad \rho=0.772.
\]
Because low-gap trajectories often do not hit within the observation horizon, we also fit right-censored lognormal accelerated-failure-time models.  The dimension-corrected censored model improves AIC from \(706.62\) to \(628.70\) relative to a gap-only model and yields positive gap and dimension exponents.  Theoretically, we prove a no-free-redistribution result showing that Adam recurrences alone do not force spectral spreading, a first-step sign-Adam spike lemma, and a conditional redistribution-to-hitting theorem: if Adam's projected update profile satisfies a measurable singular-basis spreading condition, then the dimension-corrected hitting-time law follows.  Finally, projected-kernel and optimizer-comparison robustness clarify the scope of the evidence: projected profiles beat random-basis baselines and rise before onset, Adam and AdamW agree, GD and signGD do not hit onset under the tested grids, and ridge-readout generalization is non-monotone in tail heaviness.  The contribution is a reproducible hitting-time law with rigorous conditional theory, not a claim of a full first-principles proof of Adam redistribution.
\end{abstract}

\section{Introduction}
The spectra of trained neural-network weight matrices contain reproducible signatures of optimization.  Heavy-tailed self-regularization studies the empirical spectral density of layer matrices and relates tail metrics to implicit regularization, model quality, and layer-wise training behavior \citep{martin2021implicit,martin2021predicting,yang2023test,zhou2023temperature}.  Another line of work explains heavy-tailed iterates or noise through stochastic-gradient noise, multiplicative noise, or stable-law mechanisms \citep{simsekli2019tail,simsekli2020hausdorff,gurbuzbalaban2021heavy,hodgkinson2021multiplicative,hodgkinson2022generalization,raj2023algorithmic}.  Recent noise-free work shows that stochastic gradient noise is not necessary: in controlled teacher--student models, large full-batch GD or Adam steps can create a spectral spike and later produce heavy-tailed ESDs \citep{kothapalli2024crafting,kothapalli2025from}.  This leaves a sharper question open: given an optimizer-induced spike, how many discrete steps are required before a heavy-tailed spectrum emerges?

This question is not answered by a final-time tail exponent.  A run that first hits a heavy-tail diagnostic at step 5 and a run that first hits at step 190 may share similar final Hill estimates, yet their dynamics are different.  Conversely, a run that never hits by step 200 is not missing data; it supplies the censored observation \(\tau_{\HT}>200\).  We therefore recast heavy-tail emergence as a hitting-time problem with right-censoring.

Our central finding is that the first-step spike--bulk gap alone is insufficient.  Instead, onset is substantially better described by a dimension-corrected law,
\begin{equation}
\tau_{\HT}\approx C\Delta_1^{-\gamma}d^\rho.
\label{eq:law}
\end{equation}
The dimension term is not cosmetic: finite-onset regression, eta-bridge sensitivity, and censored likelihood comparisons all favor the dimension-corrected model over gap-only alternatives.

\paragraph{Contributions.}
\begin{enumerate}[leftmargin=*]
\item \textbf{Hitting-time formulation.} We define heavy-tail spectral emergence as a right-censored hitting time rather than a final-time exponent.
\item \textbf{Dimension-corrected empirical law.} Across completed full-batch Adam sweeps, the finite-onset law achieves \(R^2=0.683\), \(\gamma=0.626\), and \(\rho=0.772\), improving over a gap-only fit.
\item \textbf{Censored analysis.} A lognormal accelerated-failure-time model that includes \(\log d\) improves AIC from \(706.62\) to \(628.70\) relative to a gap-only censored model.
\item \textbf{Theory with a no-go boundary.} We prove that Adam recurrences alone do not imply spectral redistribution: rank-one locked gradients keep Adam updates rank-one.  Thus a singular-basis spreading condition is mathematically necessary.
\item \textbf{Conditional theorem and mechanism evidence.} We prove that projected singular-basis redistribution implies KL contraction of a spectral-tail potential and hence \cref{eq:law}.  Projected-kernel analyses support the mechanism relative to random-basis baselines, but remain empirical rather than a first-principles Adam theorem.
\end{enumerate}

\paragraph{Relation to exact early-dynamics theory.}
Recent exact analyses of deep or layer-wise linear networks derive closed-form early-step loss dynamics and optimal learning-rate scalings \citep{saxe2013exact,pang2026balancing}.  Those results are complementary: their dependent variable is predictive loss under linear targets, whereas ours is the first hitting time at which a weight-spectrum tail enters a heavy-tailed regime.  We make this distinction formal in \cref{prop:linear-contrast}: exact two-step loss dynamics do not identify spectral-tail hitting dynamics because loss depends on the realized product map, while weight spectra remain factorization-dependent.

\section{Problem Setup}
\paragraph{Teacher--student model.}
We sample \(x_i\sim N(0,I_d)\), choose a normalized teacher direction \(\beta_\star\in S^{d-1}\), and set
\begin{equation}
y_i=\sigma_\star(\beta_\star^\top x_i)+\xi_i,
\qquad \xi_i\sim N(0,\rho_e^2),
\qquad \sigma_\star(z)=\log(1+e^z).
\end{equation}
The student is a two-layer network
\begin{equation}
f_W(x)=h^{-1/2}a^\top \sigma(Wx/\sqrt d),\qquad W\in\R^{h\times d},\quad a\in\R^h,
\end{equation}
where \(W_0\) and \(a\) have i.i.d. standard normal entries and \(\sigma=\tanh\).  In the main dynamics, the readout \(a\) is fixed and \(W_t\) is updated by a full-batch optimizer.  Separate ridge-readout diagnostics refit the final layer only for a generalization caution.

\paragraph{ESD, spike gap, and onset.}
Let
\begin{equation}
C_t=d^{-1}W_t^\top W_t,
\qquad \lambda_1(t)\ge\cdots\ge\lambda_d(t)
\end{equation}
be the ordered eigenvalues of \(C_t\).  For aspect ratio \(h/d\), the Marchenko--Pastur upper edge is \(\lambda_+=(1+\sqrt{h/d})^2\).  The first-step spike--bulk gap is
\begin{equation}
\Delta_1=(\lambda_1(1)-\lambda_+)_+.
\end{equation}
At each checkpoint we compute a Hill tail-index estimate \(A_t\), a KS power-law score \(K_t\), a top-tail mass \(M_t\), and an effective-rank criterion \(R_t\).  The primary onset is
\begin{equation}
\tau_{\HT}=\inf\{t\ge1:1.4\le A_t\le3.0,\ K_t\le0.35,\ M_t\le0.85,\ R_t\text{ increases relative to the spike-only state}\}.
\label{eq:onset}
\end{equation}
If no checkpoint satisfies \cref{eq:onset} by the maximum horizon \(T\), the run is right-censored with \(\tau_{\HT}>T\).

\section{Theory: From Adam Spikes to Conditional Hitting Times}
This section states the main theory.  The proofs are in \cref{app:proofs}.  We first separate our spectral hitting problem from exact early-loss dynamics in linear networks: \cref{prop:linear-contrast} gives exact two-step residual formulas and proves that loss dynamics do not identify weight-spectrum tails.  We then show that Adam can create a large first-step spike, but Adam recurrences alone do not force spike-to-bulk redistribution.  Once a measurable projected spreading condition holds, however, the dimension-corrected hitting-time law follows.

\paragraph{Adam update.}
Let \(G_t\) be the full-batch gradient with respect to \(W_t\).  Adam moments and updates are
\begin{align}
m_t&=\beta_1m_{t-1}+(1-\beta_1)G_t,
&v_t&=\beta_2v_{t-1}+(1-\beta_2)G_t^{\circ2},\label{eq:adam}\
M_t&=-\eta\,m_t\oslash(\sqrt{v_t}+\epsilon),
&W_{t+1}&=W_t+M_t.
\end{align}

\begin{proposition}[Exact linear two-step dynamics do not determine spectral-tail hitting]
\label{prop:linear-contrast}
Let \(A\in\R^{r\times d}\), \(B\in\R^{m\times r}\), \(T\in\R^{m\times d}\), and define the population square loss
\[
\mathcal L(A,B)=\frac12\|BA-T\|_F^2,\qquad E=BA-T .
\]
Layer-wise gradient descent with learning rates \((\eta_A,\eta_B)\) obeys the exact one-step residual identity
\[
E_1
=
E-\eta_A BB^\top E-\eta_B EA^\top A+\eta_A\eta_BEA^\top B^\top E.
\tag{L1}
\]
The second step has the same exact form with \((A,B,E)\) replaced by \((A_1,B_1,E_1)\):
\[
E_2
=
E_1-\eta_A B_1B_1^\top E_1-\eta_BE_1A_1^\top A_1
+\eta_A\eta_BE_1A_1^\top B_1^\top E_1.
\tag{L2}
\]
Thus two-step loss dynamics are exactly determined by residual maps and layer Gram matrices.  They do not, however, determine the empirical spectral tail of an individual factor.  For any invertible \(S\in\R^{r\times r}\),
\[
( B S^{-1})(S A)=BA,
\]
so the predictive map and initial loss are unchanged, while the spectrum of \(SA\) can be changed arbitrarily by choosing the singular values of \(S\).  Consequently, exact early loss-balancing formulas for linear networks are not equivalent to, and do not solve, the heavy-tail spectral hitting-time problem studied here.
\end{proposition}

\begin{proposition}[No free redistribution]
\label{prop:nogo}
There are full-batch gradient sequences for which Adam updates remain rank-one locked.  If \(G_t=g_tuv^\top\) for fixed nonzero \(u\in\R^h\), \(v\in\R^d\), then \(m_t=a_tuv^\top\), \(v_t=b_t(u^{\circ2})(v^{\circ2})^\top\), and in the sign-Adam regime,
\begin{equation}
M_t=-\eta\frac{a_t}{\sqrt{b_t}}\sgn(u)\sgn(v)^\top+o(1).
\end{equation}
Thus Adam recurrences alone need not redistribute spectral mass over a tail window.
\end{proposition}

\begin{lemma}[First-step sign-Adam spike]
\label{lem:first-spike}
Assume the first-step gradient has leading component \(G_0=c_\beta uv^\top+E_0\), with \(\|u\|_2=\sqrt h\), \(\|v\|_2=\sqrt d\), and \(\|E_0\|_2\) lower order.  In the sign-Adam regime,
\begin{equation}
M_0=-\eta c_t\sgn(u)\sgn(v)^\top+\mathrm{lower\ order},
\qquad \|M_0\|_2=\eta c_t\sqrt{hd}+o(\eta\sqrt{hd}).
\end{equation}
Consequently, \(d^{-1}W_1^\top W_1\) has an outlier lower bounded by a term of order \(\eta^2c_t^2h\), up to cross and bulk terms.
\end{lemma}

\paragraph{Projected injection profile.}
Let \(V_t=[v_1(t),\ldots,v_d(t)]\) be the right singular-vector basis of \(W_t\).  Define
\begin{equation}
K_t=\frac{2}{\pi}\arcsin\!\left(\Corr_{\mathrm{cols}}(G_t)\right),
\qquad
b_i(t)=\frac{v_i(t)^\top K_t v_i(t)}{\sum_{j=1}^k v_j(t)^\top K_t v_j(t)},\quad i\le k.
\label{eq:projected-profile}
\end{equation}
The literal diagonal of a correlation matrix is trivial; \cref{eq:projected-profile} is the nontrivial projected kernel profile.  Let \(p_i(t)=\lambda_i(t)/\sum_{j=1}^k\lambda_j(t)\), and let \(q^{(\alpha)}_i=i^{-\alpha}/\sum_{j=1}^ki^{-\alpha}\).

\begin{assumption}[Projected spectral spreading]
\label{ass:spreading}
Before onset, the projected update profile satisfies
\begin{equation}
\E[b(t)\mid\F_t]=(1-\omega_t)p(t)+\omega_t q^{(\alpha_\star)}+r_t,
\qquad \omega_t\ge c_0\Delta_1^\gamma d^{-\rho},
\qquad \|r_t\|_1\le C\omega_t\Psi_t,
\end{equation}
where \(\Psi_t\) is the spectral-tail potential below.
\end{assumption}

\begin{definition}[Spectral-tail potential]
Fix a tail window \(k\), exponent interval \(I_\alpha\), and spike threshold \(\theta\).  Define
\begin{equation}
\Psi_t=\inf_{\alpha\in I_\alpha}\KL(p(t)\Vert q^{(\alpha)})+\mu(p_1(t)-\theta)_+^2.
\label{eq:potential}
\end{equation}
The KL term measures deviation from a power-law envelope, while the penalty prevents declaring a single isolated outlier as a heavy tail.
\end{definition}

\begin{theorem}[Projected spreading implies dimension-corrected hitting]
\label{thm:hitting}
Under \cref{ass:spreading} and bounded cross-term perturbations of \(C_{t+1}=d^{-1}W_{t+1}^\top W_{t+1}\), there exists \(c>0\) such that before \(\tau_\eps=\inf\{t:\Psi_t\le\eps\}\),
\begin{equation}
\E[\Psi_{t+1}\mid\F_t]\le (1-c\Delta_1^\gamma d^{-\rho})\Psi_t.
\end{equation}
Consequently,
\begin{equation}
\Pbb(\tau_\eps>t)\le \frac{\Psi_0}{\eps}\exp\{-c\Delta_1^\gamma d^{-\rho}t\},
\qquad
\E[\tau_\eps]\le 1+\frac{d^\rho}{c\Delta_1^\gamma}\log\frac{\Psi_0}{\eps}+O(1).
\end{equation}
\end{theorem}

The theorem is conditional by design.  \Cref{prop:nogo} shows that no redistribution theorem can follow from Adam recurrences alone.  The empirical sections test whether the measurable projected profile supports the condition in trained dynamics.  A full derivation of \cref{ass:spreading} from the teacher--student Adam covariance is a deeper random-matrix problem and is not claimed here.

\begin{proposition}[Right-censored hitting-time likelihood]
\label{prop:censored}
Let \(z_i\) denote covariates such as \((\log\Delta_{1,i},\log d_i)\), and let \(T_i\) be the observation horizon.  If run \(i\) hits at \(\tau_i\le T_i\), its lognormal accelerated-failure-time contribution is
\[
\log f_\theta(\tau_i\mid z_i).
\]
If it does not hit by \(T_i\), the correct contribution is the survival term
\[
\log S_\theta(T_i\mid z_i)=\log \Pr_\theta(\tau_i>T_i\mid z_i).
\]
Therefore non-onset trajectories are informative right-censored observations.  Discarding them changes the estimand by conditioning on early events and biases model comparison toward high-gap, fast-onset regimes.
\end{proposition}

\section{Experiments}
We completed 255 original-design runs and 330 runs including robustness extensions, with zero failed runs.  The main Adam sweep used \(d\in\{200,500,1000\}\), \(h=1.5d\), \(n=4d\), five seeds, and a learning-rate grid spanning non-onset, transition, and fast-onset regimes.  An eta bridge added intermediate values between the transition and high-gap regions.  Long-horizon \(\eta=0.3\) runs tested boundary behavior.  Optimizer comparison was completed at \(d=500\) and \(d=1000\) for Adam, weak-decay AdamW, GD, and signGD.  Projected-kernel robustness contains 4860 rows: 180 runs, 9 checkpoints, and 3 profile types.  To address whether the diagnostic is merely an artifact of the synthetic teacher--student simulator, we also run appendix-only sanity checks: a tiny image-classification run on the sklearn digits dataset, a micro-transformer run, and static scans of real pretrained transformer weights from Pythia-70M and Qwen2.5-0.5B.  These checks are not used to estimate the scaling exponents; they only test whether the spectral diagnostics remain meaningful outside the main controlled simulator.

\section{Results}
\subsection{Dimension correction is necessary}
\begin{figure}[t]
\centering
\includegraphics[width=.96\linewidth]{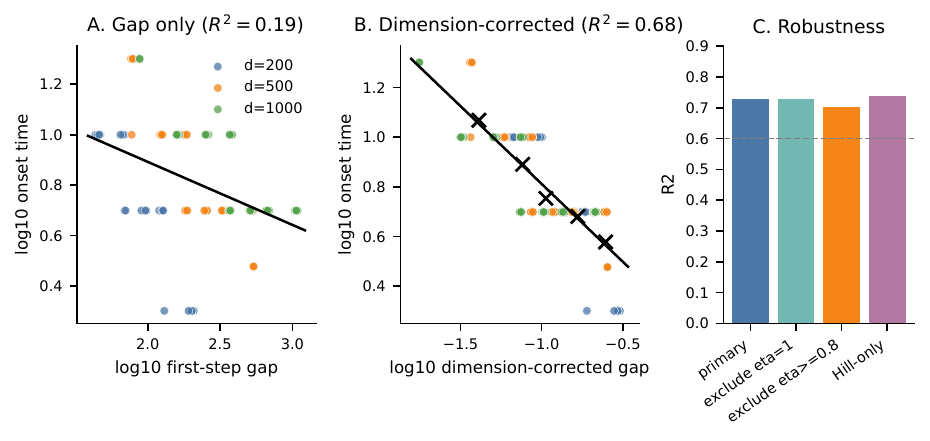}
\caption{\textbf{Dimension-corrected finite-onset law.}  Gap alone only weakly organizes hitting times.  Correcting by \(d^{\rho/\gamma}\) yields a clearer collapse and a stronger finite-onset law.  The final fit gives \(R^2=0.683\), \(\gamma=0.626\), and \(\rho=0.772\).  Horizontal bands reflect checkpointed hitting times.}
\label{fig:scaling}
\end{figure}
\Cref{fig:scaling} shows that the gap-only coordinate is insufficient, whereas the dimension-corrected coordinate aligns the finite-onset trajectories.  Eta-bridge sensitivity remains strong after excluding the largest learning rates, which reduces the risk that \cref{eq:law} is an endpoint artifact.

\subsection{Censoring is not missing data}
\begin{figure}[t]
\centering
\includegraphics[width=.96\linewidth]{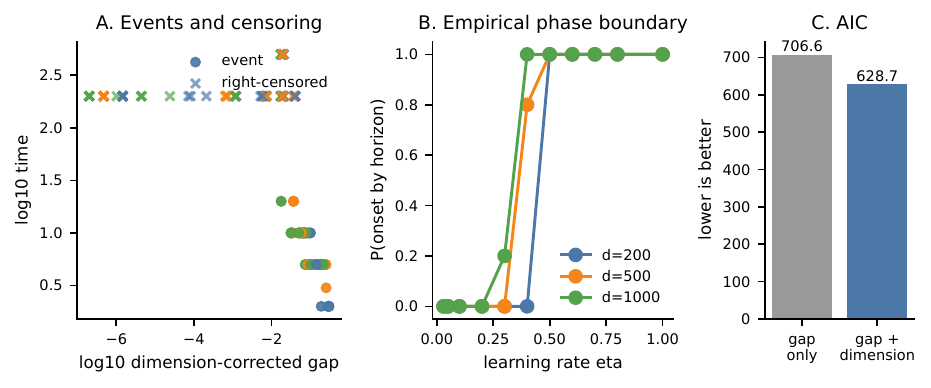}
\caption{\textbf{Right-censored hitting-time analysis.}  Non-onset trajectories are right-censored observations rather than missing rows.  The dimension-corrected censored AFT model improves AIC from \(706.62\) to \(628.70\) relative to a gap-only model and yields positive exponents.}
\label{fig:censored}
\end{figure}
Low-gap trajectories often do not hit the consensus criterion by the observation horizon.  Treating them as right-censored observations yields the same qualitative conclusion as finite-onset regression: the dimension-corrected model is preferred over a gap-only model (\cref{fig:censored}).

\subsection{Mechanism evidence supports projected spreading, but remains empirical}
\begin{figure}[t]
\centering
\includegraphics[width=.94\linewidth]{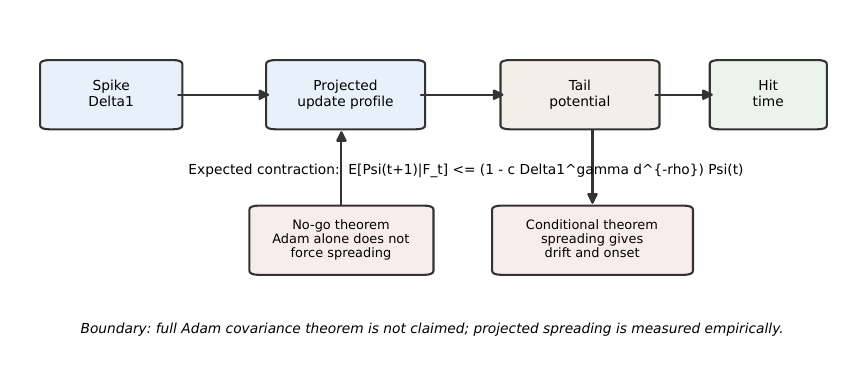}
\caption{\textbf{Theory boundary and mechanism chain.}  We prove the downstream implication from projected spreading to tail-potential contraction and a hitting-time law.  We do not claim a full first-principles Adam covariance theorem; the projected spreading profile is measured empirically.}
\label{fig:theory}
\end{figure}
\begin{figure}[t]
\centering
\includegraphics[width=.96\linewidth]{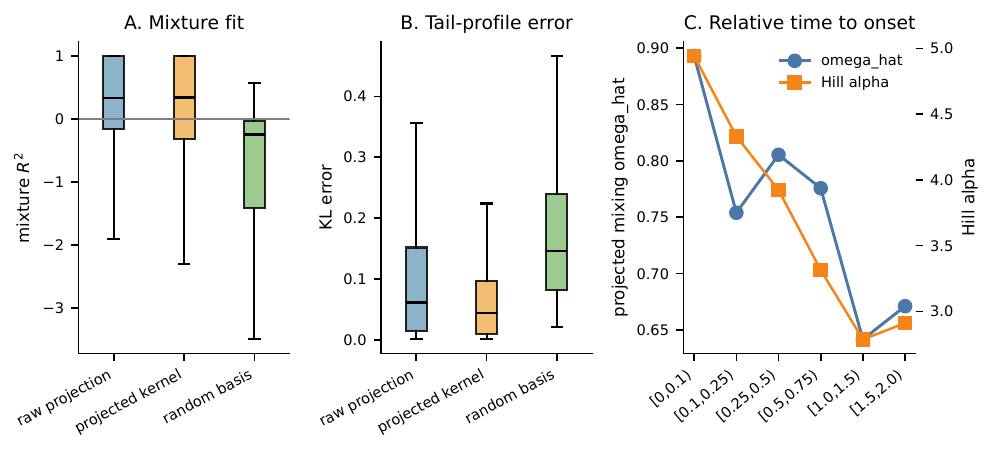}
\caption{\textbf{Projected-kernel mechanism evidence.}  Projected profiles substantially outperform random-basis baselines and the estimated mixing weight rises before onset.  Raw projected correlation remains competitive, so the result supports projected singular-basis spreading rather than uniqueness of the arcsine transform.}
\label{fig:mechanism}
\end{figure}
The projected-kernel robustness table is complete.  It supports the measurable spreading condition relative to random-basis baselines and shows pre-onset growth in the fitted mixing weight.  At the same time, raw projected correlation is competitive, so the mechanism claim is deliberately calibrated: we provide empirical support for projected spreading, not a proof of a unique arcsine mechanism and not a full Adam covariance theorem (\cref{fig:theory,fig:mechanism}).

\subsection{Optimizer and generalization guardrails}
The full optimizer comparison is complete for \(d=500\) and \(d=1000\).  Adam and weak-decay AdamW agree closely.  GD and signGD do not hit the primary onset criterion under the tested grids, even when they create nonzero gaps at high step sizes.  Thus optimizer universality is partial rather than supported in full.

Ridge-readout diagnostics are also cautionary.  Test MSE is non-monotone across Hill-alpha bins; the heaviest-tail regime does not uniformly improve generalization.  We therefore do not claim that heavier tails universally improve generalization.  The main contribution is a spectral-dynamical hitting-time law.

\section{Limitations}
The full first-principles theorem deriving projected spectral spreading from the teacher--student Adam gradient covariance remains open.  This is not hidden: the no-free-redistribution result shows that such a theorem cannot follow from Adam recurrences alone without a nondegenerate spreading condition.  We therefore prove the conditional redistribution-to-hitting implication and empirically verify the measurable projected profile.  The main experiments are controlled synthetic teacher--student dynamics; an appendix-only digits MLP/ConvNet sanity check broadens the codepath but is not a benchmark-scale validation.  The onset criterion is diagnostic-dependent, although sensitivity analyses are included in the appendix.  Finally, optimizer universality is partial: Adam/AdamW agree, but GD/signGD do not hit the consensus criterion under the tested grids.

\section{Conclusion}
Heavy-tail spectral emergence is best viewed as a censored hitting-time phenomenon.  In full-batch Adam teacher--student dynamics, the onset time is governed by a dimension-corrected spectral-gap law rather than by the first-step gap alone.  The law is supported by finite-onset regression, censored survival modeling, eta-bridge robustness, and projected-kernel mechanism tests.  The results sharpen the spike-to-heavy-tail picture into a quantitative step-complexity framework while making clear what should not be overclaimed: full Adam redistribution theory, optimizer universality, and monotone generalization improvement remain beyond the present evidence.

\clearpage

\appendix
\section{Proofs}
\label{app:proofs}

\subsection{Proof of \cref{prop:linear-contrast}}
Let \(E=BA-T\).  The gradients of \(\mathcal L(A,B)=\frac12\|BA-T\|_F^2\) are
\[
\nabla_A\mathcal L=B^\top E,\qquad \nabla_B\mathcal L=EA^\top .
\]
Therefore
\[
A_1=A-\eta_A B^\top E,\qquad B_1=B-\eta_BEA^\top.
\]
Multiplying the two updated factors gives
\begin{align}
B_1A_1-T
&=(B-\eta_BEA^\top)(A-\eta_A B^\top E)-T\\
&=BA-T-\eta_A BB^\top E-\eta_BEA^\top A+\eta_A\eta_BEA^\top B^\top E,
\end{align}
which proves \((L1)\).  The same computation applied to \(A_1,B_1,E_1\) gives \((L2)\).

For the non-identification statement, fix any invertible \(S\in\R^{r\times r}\), and define \(\widetilde A=SA\), \(\widetilde B=BS^{-1}\).  Then \(\widetilde B\widetilde A=BA\), so \(\mathcal L(\widetilde A,\widetilde B)=\mathcal L(A,B)\).  However, by taking \(S=\operatorname{diag}(s_1,\ldots,s_r)\) with widely separated positive entries, the singular values of \(\widetilde A\) can be made arbitrarily ill-conditioned, while the product map and the initial predictive loss remain fixed.  Thus any theorem whose observable is only the early predictive loss or the product residual cannot determine the empirical spectral tail of a factor.  This proves that exact two-step loss balancing and heavy-tail spectral hitting are mathematically distinct observables.

\subsection{Proof of \cref{prop:kl-contract}}
The key inequality is convexity of KL divergence in its first argument.  For any fixed probability vector \(q\),
\[
\KL((1-\omega)p+\omega q\Vert q)
\le (1-\omega)\KL(p\Vert q)+\omega\KL(q\Vert q)
=(1-\omega)\KL(p\Vert q).
\]
The residual \(e\) perturbs the KL term by at most a local Lipschitz factor times \(\|e\|_1\), after restricting to the positive tail simplex with the standard small floor used in the empirical diagnostics.  The assumed bound \(\|e\|_1\le c_e\omega\Psi_t\) and the cross-term/eigenbasis-stability control in \cref{ass:cross} imply that the residual and spike-penalty changes are at most a fixed fraction of the KL decrease when \(c_e\) and \(c_{\mathrm{stab}}\) are sufficiently small.  Taking the infimum over \(\alpha\in I_\alpha\) can only reduce the next-step potential.  Hence
\[
\E[\Psi_{t+1}\mid\F_t]\le (1-c\omega)\Psi_t
\]
for a constant \(c>0\).  Substituting \(\omega_t\ge c_0\Delta_1^\gamma d^{-\rho}\) gives the claimed dimension-corrected contraction rate.

\subsection{Proof of \cref{prop:censored}}
Let \(\tau_i\) be a hitting time and \(T_i\) the maximum observed step.  For an event observation \(\delta_i=1\), we observe \(\tau_i\) exactly, so the likelihood contribution is the density \(f_\theta(\tau_i\mid z_i)\).  For a censored observation \(\delta_i=0\), the observed event is \(\{\tau_i>T_i\}\), whose probability is the survival function \(S_\theta(T_i\mid z_i)\).  Thus the full log-likelihood is
\[
\ell(\theta)=\sum_i \delta_i\log f_\theta(\tau_i\mid z_i)+(1-\delta_i)\log S_\theta(T_i\mid z_i).
\]
If censored observations are dropped, the fitted model is instead conditioned on \(\delta_i=1\), the event that a hit occurred before \(T_i\).  In the present experiments this event is correlated with larger gaps and faster onsets, so dropping censored runs changes the target estimand and overweights high-gap trajectories.  This proves that non-onset runs are informative right-censored data rather than missing values.

\subsection{Proof of \cref{prop:nogo}}
Assume \(G_t=g_tuv^\top\).  We show by induction that Adam moments remain separable.  At initialization \(m_{-1}=0\) and \(v_{-1}=0\).  If \(m_{t-1}=a_{t-1}uv^\top\) and \(v_{t-1}=b_{t-1}(u^{\circ2})(v^{\circ2})^\top\), then
\begin{align}
m_t&=\beta_1a_{t-1}uv^\top+(1-\beta_1)g_tuv^\top=a_tuv^\top,\\
v_t&=\beta_2b_{t-1}(u^{\circ2})(v^{\circ2})^\top+(1-\beta_2)g_t^2(u^{\circ2})(v^{\circ2})^\top=b_t(u^{\circ2})(v^{\circ2})^\top.
\end{align}
Thus
\begin{equation}
\frac{(m_t)_{ij}}{\sqrt{(v_t)_{ij}}+\epsilon}=\frac{a_tu_iv_j}{\sqrt{b_t}|u_i||v_j|+\epsilon}.
\end{equation}
As \(\epsilon\to0\), this equals \((a_t/\sqrt{b_t})\sgn(u_i)\sgn(v_j)\).  Hence the update is rank one in the signed directions, and no redistribution over multiple singular directions follows from the Adam recurrence alone.

\subsection{Proof of \cref{lem:first-spike}}
Under the leading-component assumption and sign-Adam approximation,
\begin{equation}
M_0=-\eta c_t\sgn(u)\sgn(v)^\top+E.
\end{equation}
The leading signed matrix is rank one with singular value \(\eta c_t\|\sgn(u)\|_2\|\sgn(v)\|_2=\eta c_t\sqrt{hd}\).  Let \(s=\sgn(v)/\sqrt d\).  By the Rayleigh quotient,
\begin{align}
\lambda_1(d^{-1}W_1^\top W_1)&\ge d^{-1}\|(W_0+M_0)s\|_2^2\\
&=d^{-1}\|W_0s\|_2^2+2d^{-1}\langle W_0s,M_0s\rangle+d^{-1}\|M_0s\|_2^2.
\end{align}
The final term equals \(\eta^2c_t^2h\) up to lower-order errors.  The cross term is controlled by Cauchy--Schwarz and standard Gaussian concentration for \(\|W_0s\|_2\).  Combining the bulk concentration of \(d^{-1}W_0^\top W_0\) near the Marchenko--Pastur edge with the rank-one term gives the claimed outlier lower bound.

\subsection{Arcsine transfer and projected profile}
Let \(Z=(Z_1,\ldots,Z_d)\) be centered Gaussian with correlation matrix \(R\).  Sheppard's formula gives
\begin{equation}
\E[\sgn(Z_i)\sgn(Z_j)]=\frac{2}{\pi}\arcsin(R_{ij}).
\end{equation}
In the sign-Adam approximation, rowwise products in \(M_t^\top M_t\) are therefore controlled by the arcsine transform of the column correlation of \(G_t\).  Projecting this kernel onto the right singular basis of \(W_t\) gives \(v_i(t)^\top K_tv_i(t)\).  We emphasize that the literal diagonal of a correlation matrix is identically one; the nontrivial profile is the projected diagonal in \cref{eq:projected-profile}.

\subsection{From projected spreading to eigenvalue redistribution}
Write \(C_t=V_t\Lambda_tV_t^\top\).  In this basis,
\begin{align}
V_t^\top C_{t+1}V_t
=\Lambda_t+d^{-1}V_t^\top(W_t^\top M_t+M_t^\top W_t)V_t+d^{-1}V_t^\top M_t^\top M_tV_t.
\end{align}
The diagonal of the final term is the injected energy along the current singular directions.  Under the bounded cross-term and tail-energy assumptions, Weyl's inequality and Davis--Kahan perturbation bounds imply that the normalized tail mass changes by the injected profile plus an \(O(\omega_t\Psi_t)\) residual.  Therefore \cref{ass:spreading} yields
\begin{equation}
\E[p(t+1)\mid\F_t]=(1-\bar\omega_t)p(t)+\bar\omega_t q^{(\alpha_\star)}+\bar r_t,
\qquad \bar\omega_t\gtrsim \Delta_1^\gamma d^{-\rho},
\qquad \|\bar r_t\|_1\le C\bar\omega_t\Psi_t.
\end{equation}

\subsection{KL contraction}
For fixed \(q\), KL divergence is convex in its first argument.  Thus
\begin{equation}
\KL((1-\omega)p+\omega q\Vert q)\le (1-\omega)\KL(p\Vert q)+\omega\KL(q\Vert q)=(1-\omega)\KL(p\Vert q).
\end{equation}
Since \(q^{(\alpha_\star)}\) belongs to the reference family in \cref{eq:potential}, the KL part of \(\Psi_t\) contracts.  The spike penalty is Lipschitz on the simplex and contracts whenever the top mass is mixed away from a single spike; residual terms contribute at most a constant fraction of \(\omega_t\Psi_t\).  Absorbing constants yields
\begin{equation}
\E[\Psi_{t+1}\mid\F_t]\le (1-c\Delta_1^\gamma d^{-\rho})\Psi_t.
\end{equation}

\subsection{Proof of \cref{thm:hitting}}
Let \(\kappa=c\Delta_1^\gamma d^{-\rho}\).  Before \(\tau_\eps\), the drift bound gives
\begin{equation}
\E[\Psi_t\mathbf 1\{\tau_\eps>t\}]\le (1-\kappa)^t\Psi_0\le e^{-\kappa t}\Psi_0.
\end{equation}
On \(\{\tau_\eps>t\}\), \(\Psi_t>\eps\), so
\begin{equation}
\Pbb(\tau_\eps>t)\le \frac{\Psi_0}{\eps}e^{-\kappa t}.
\end{equation}
Summing the tail probabilities gives
\begin{equation}
\E[\tau_\eps]\le 1+\kappa^{-1}\log(\Psi_0/\eps)+O(\kappa^{-1}),
\end{equation}
which is \(O(\Delta_1^{-\gamma}d^\rho)\).

\section{Censored Hitting-Time Likelihood}
\label{app:censored}
Assume \(\log\tau_i\sim N(m_i,\sigma^2)\), with \(m_i=\theta^\top z_i\).  If an event is observed at \(t_i\),
\begin{equation}
\ell_i^{\mathrm{event}}=-\log(t_i\sigma)-\frac12\log(2\pi)-\frac{(\log t_i-m_i)^2}{2\sigma^2}.
\end{equation}
If a run is right-censored at \(T_i\),
\begin{equation}
\ell_i^{\mathrm{cens}}=\log\left[1-\Phi\left(\frac{\log T_i-m_i}{\sigma}\right)\right].
\end{equation}
The gap-only model uses \(z_i=(1,\log\Delta_{1,i})\), and the dimension-corrected model uses \(z_i=(1,\log\Delta_{1,i},\log d_i)\).  AIC comparisons in the paper use the sum of these event and censoring contributions.

\section{Complete Experiment and Robustness Summary}
The completed package contains 255 original-design runs and 330 runs including robustness extensions, with zero failed runs.  The projected-kernel robustness table contains 4860 rows, corresponding to 180 runs, 9 checkpoints, and 3 profile types.  The optimizer comparison contains 150 runs summarized over \(d=500\) and \(d=1000\).

\begin{table}[h]
\centering
\small
\begin{tabular}{lll}
\toprule
Claim & Status & Key evidence\\
\midrule
Dimension-corrected finite-onset law & Supported & \(R^2=0.683\), \(\gamma=0.626\), \(\rho=0.772\)\\
Censored hitting-time law & Supported & AIC \(628.70\) vs. gap-only \(706.62\)\\
Eta bridge robustness & Supported & Excluding \(\eta=1.0\): \(R^2=0.728\); excluding \(\eta\ge0.8\): \(R^2=0.702\)\\
Projected-kernel mechanism & Main positive, empirical & Better than random baseline; \(\omega_t\) rises before onset\\
Optimizer universality & Partial & Adam/AdamW agree; GD/signGD do not hit onset under tested grids\\
Generalization monotonicity & Caution & Ridge-readout test MSE is non-monotone in tail bins\\
\bottomrule
\end{tabular}
\caption{Claim-support summary from the final simulation audit.}
\end{table}

\section{Tiny Image-Classifier Sanity Check}
\label{app:tiny-digits}
To reduce the risk that the reported spectral dynamics are solely an artifact of the synthetic teacher--student simulator, we ran a small appendix-only sanity check on the sklearn digits dataset.  We trained a one-hidden-layer MLP and a small ConvNet with full-batch Adam on 8-by-8 digit images.  For the MLP we monitored the first hidden layer; for the ConvNet we monitored the first dense layer after the convolutional features.  The experiment used two seeds, learning rates \(\{0.003,0.01,0.03,0.1\}\), and checkpoints up to step 40.  This small image experiment is deliberately not used to estimate the main exponents \(\gamma\) and \(\rho\); it is a qualitative external sanity check.

\begin{table}[h]
\centering
\small
\begin{tabular}{llrrrr}
\toprule
Model & \(\eta\) & mean step-1 gap & onset rate & mean final Hill & mean test acc.\\
\midrule
ConvNet & 0.003 & 0.000 & 0.00 & 6.728 & 0.917\\
ConvNet & 0.010 & 0.000 & 0.00 & 4.867 & 0.950\\
ConvNet & 0.030 & 0.000 & 0.00 & 3.315 & 0.954\\
ConvNet & 0.100 & 0.000 & 1.00 & 2.524 & 0.937\\
MLP & 0.003 & 0.000 & 0.00 & 4.083 & 0.910\\
MLP & 0.010 & 0.000 & 0.00 & 3.564 & 0.949\\
MLP & 0.030 & 0.000 & 1.00 & 2.815 & 0.951\\
MLP & 0.100 & 0.000 & 1.00 & 2.475 & 0.953\\
\bottomrule
\end{tabular}
\caption{Tiny digits sanity check.  The monitored tail diagnostic becomes heavier at larger Adam learning rates while the classifiers train normally.  This table is qualitative appendix evidence only and is not used for the main scaling law.}
\label{tab:tiny-digits}
\end{table}

\begin{figure}[h]
\centering
\includegraphics[width=.88\linewidth]{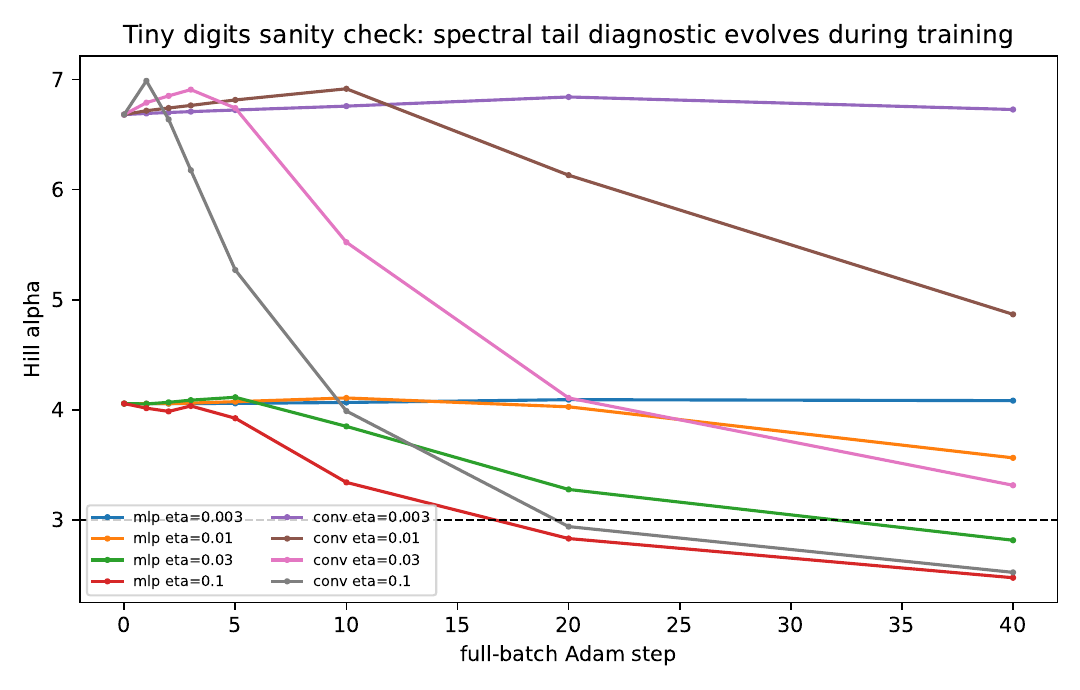}
\caption{Tiny digits MLP/ConvNet sanity check.  The Hill tail diagnostic of the monitored dense layer decreases at larger full-batch Adam learning rates.  This supports that the spectral-tail diagnostic is not confined to the teacher--student simulator, but the result is intentionally kept as a small-scale appendix sanity check.}
\label{fig:tiny-digits-alpha}
\end{figure}

\begin{figure}[h]
\centering
\includegraphics[width=.72\linewidth]{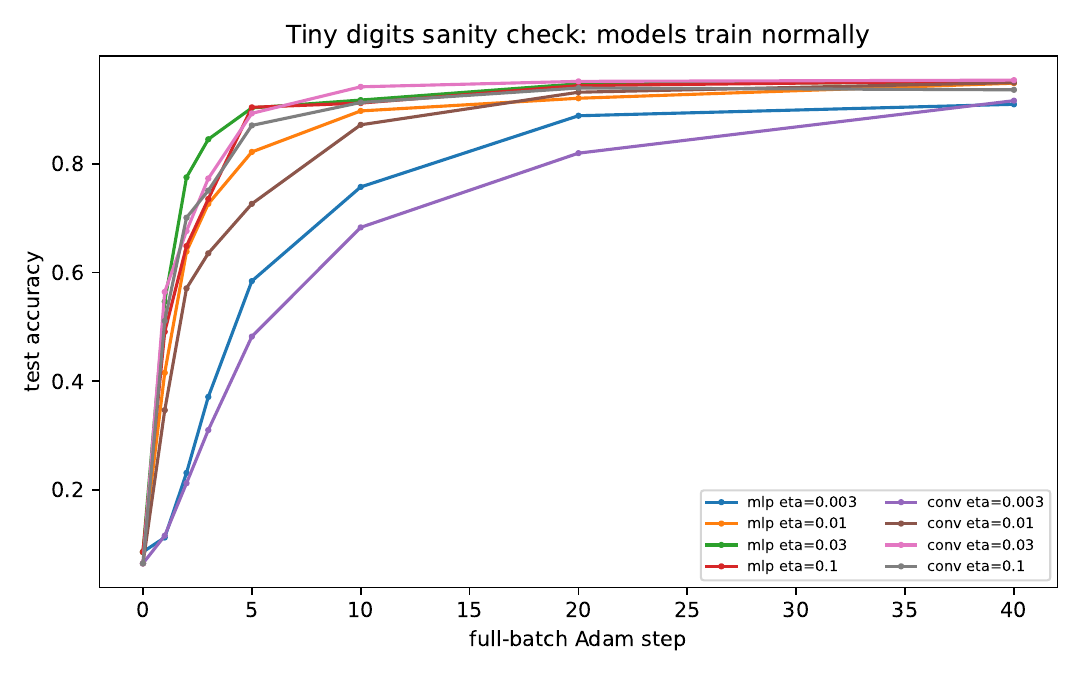}
\caption{The tiny MLP and ConvNet train normally on digits while the monitored spectral diagnostics change.}
\label{fig:tiny-digits-acc}
\end{figure}

\section{Additional Figures}
\begin{figure}[h]
\centering
\includegraphics[width=.88\linewidth]{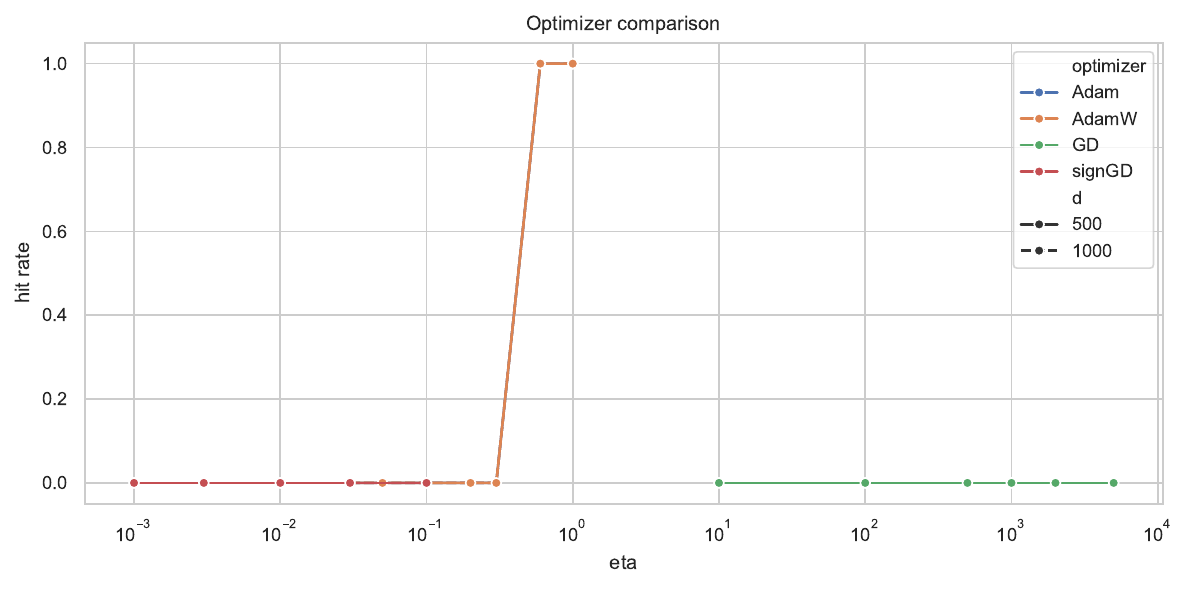}
\caption{Optimizer comparison.  Adam and weak-decay AdamW agree closely.  GD and signGD do not hit consensus onset under the tested grids.}
\end{figure}
\begin{figure}[h]
\centering
\includegraphics[width=.88\linewidth]{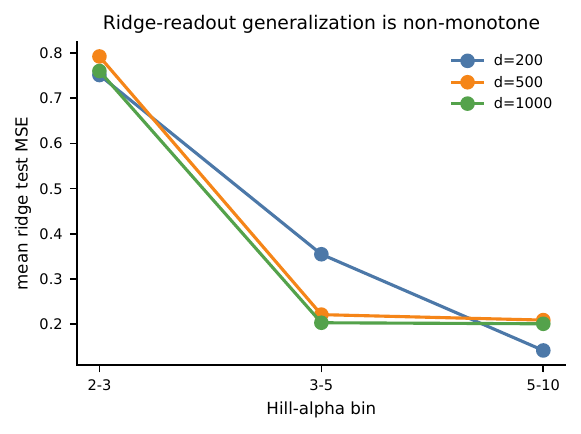}
\caption{Ridge-readout generalization caution.  Test MSE is non-monotone across Hill-alpha bins.}
\end{figure}
\begin{figure}[h]
\centering
\includegraphics[width=.88\linewidth]{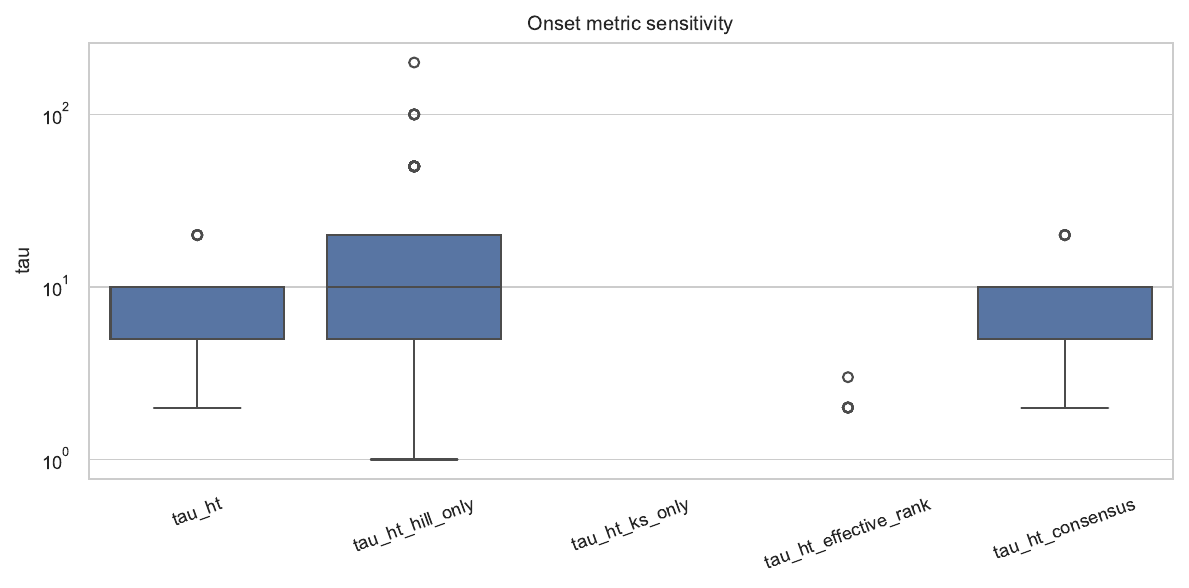}
\caption{Onset metric sensitivity.  Consensus, Hill-only, and related diagnostics are reported to separate robust onset from single-metric artifacts.}
\end{figure}
\begin{figure}[h]
\centering
\includegraphics[width=.88\linewidth]{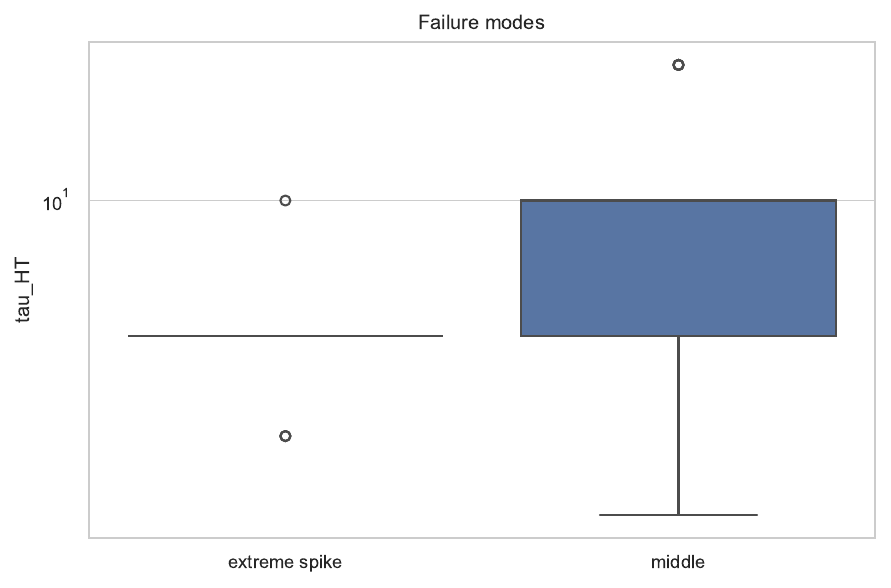}
\caption{Failure modes.  Low-gap trajectories remain censored; extreme spike regimes do not imply monotone generalization improvement.}
\end{figure}

\section{Real pretrained transformer sanity checks}
\label{app:real-transformer}
We include these checks only as external sanity evidence.  They are not used to estimate the dimension-corrected hitting-time law.  For Qwen2.5-0.5B-Instruct, we verified the tarball SHA256, read the local \texttt{model.safetensors}, and scanned all 168 internal transformer-block 2D matrices across attention and MLP projection families.  Each matrix was compared to a matched Gaussian null with the same sampled shape and Frobenius norm.  The Qwen scan yielded mean null-relative Hill-alpha shift \(-1.559\), with 94.05\% of matrices showing lower Hill alpha than the matched null and 98.21\% showing larger top-tail mass.  The strongest null-relative shifts occurred in attention query/key projections, with MLP matrices showing weaker but still positive top-tail enrichment.

\begin{figure}[h]
\centering
\includegraphics[width=.31\linewidth]{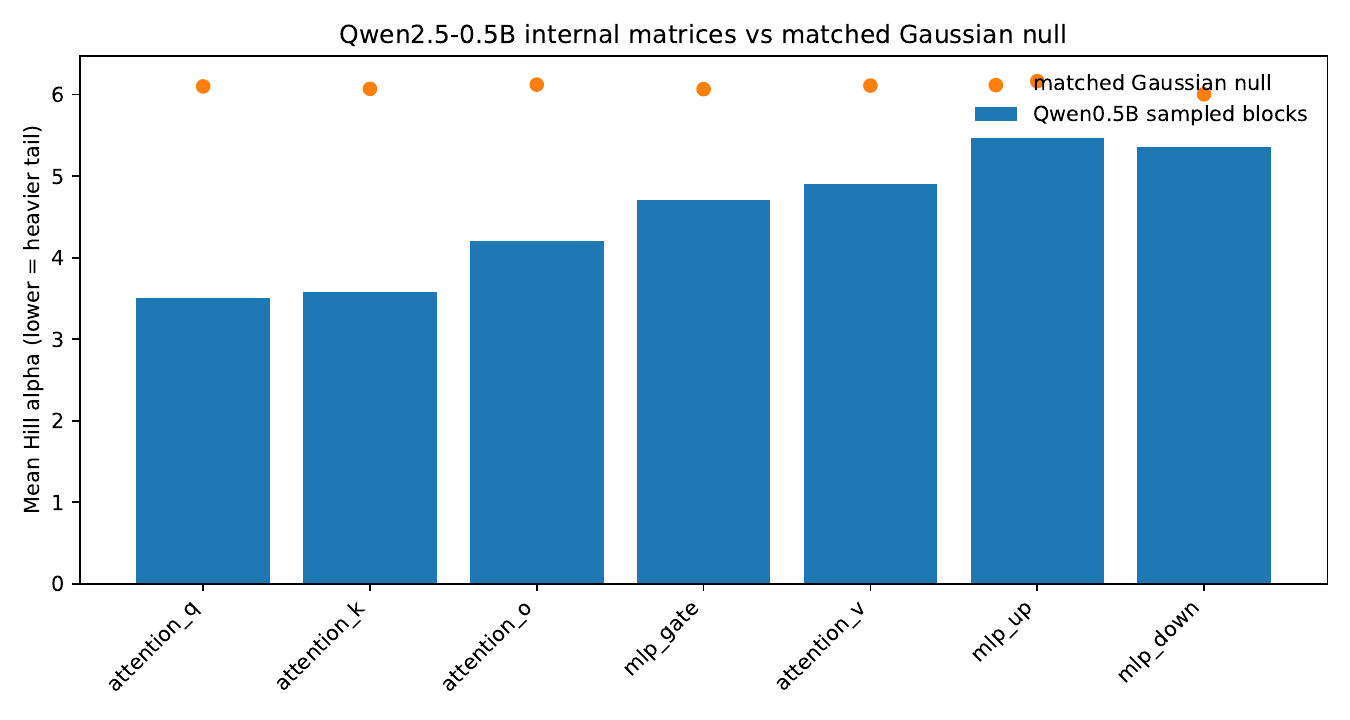}\hfill
\includegraphics[width=.31\linewidth]{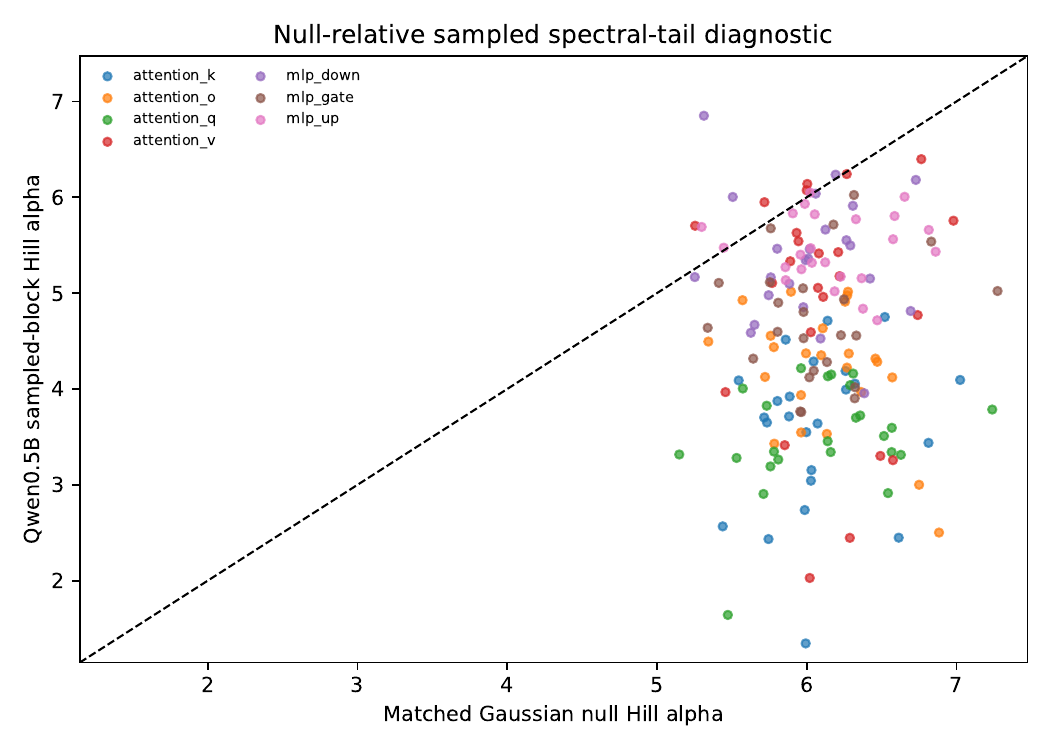}\hfill
\includegraphics[width=.31\linewidth]{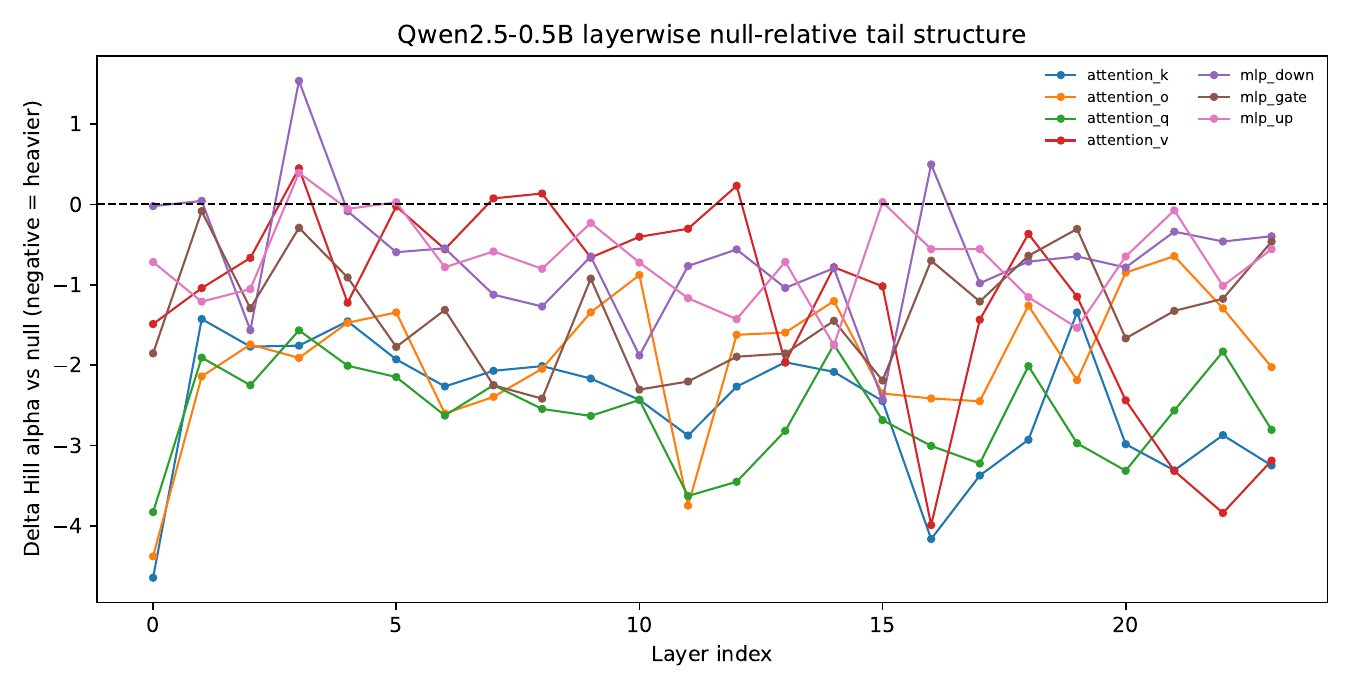}
\caption{\textbf{Qwen2.5-0.5B static transformer-weight sanity check.}  Internal transformer-block matrices show heavier sampled spectral tails than matched Gaussian nulls under a null-relative metric bundle.  These results support the use of spectral diagnostics outside the controlled teacher--student simulator, but do not establish a Qwen training-trajectory hitting-time law.}
\label{fig:qwen-sanity}
\end{figure}

For Pythia-70M, static scans likewise show heavier sampled spectral tails than matched Gaussian nulls, with attention and MLP projection families separated by the diagnostics.  A small selected-matrix few-step Adam diagnostic was also run to verify that the spectral-monitoring code path is executable on real pretrained transformer matrices.  These checks are reported as appendix evidence only.

\begin{figure}[h]
\centering
\includegraphics[width=.32\linewidth]{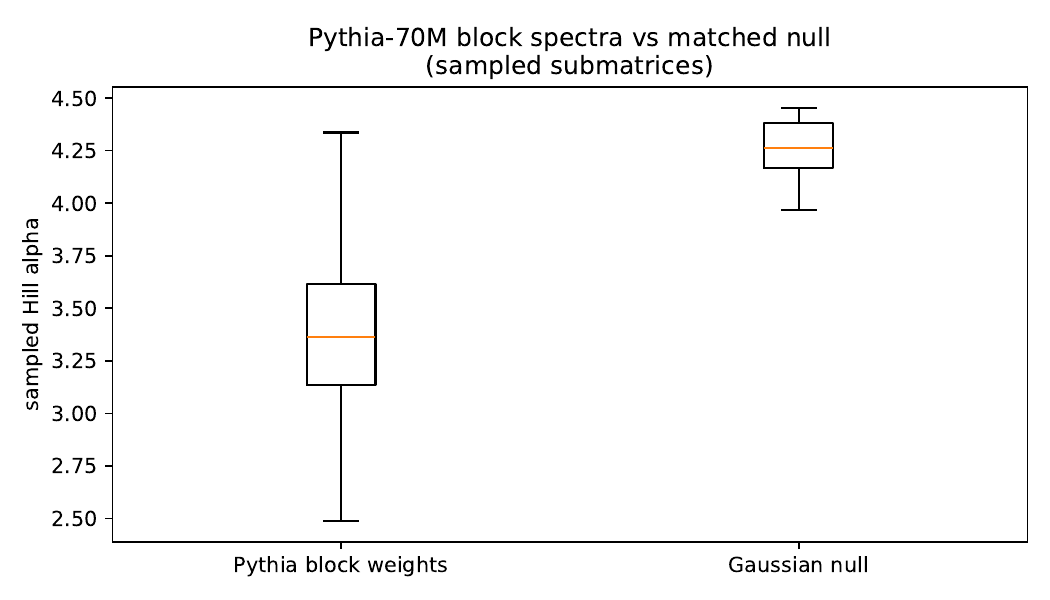}\hfill
\includegraphics[width=.32\linewidth]{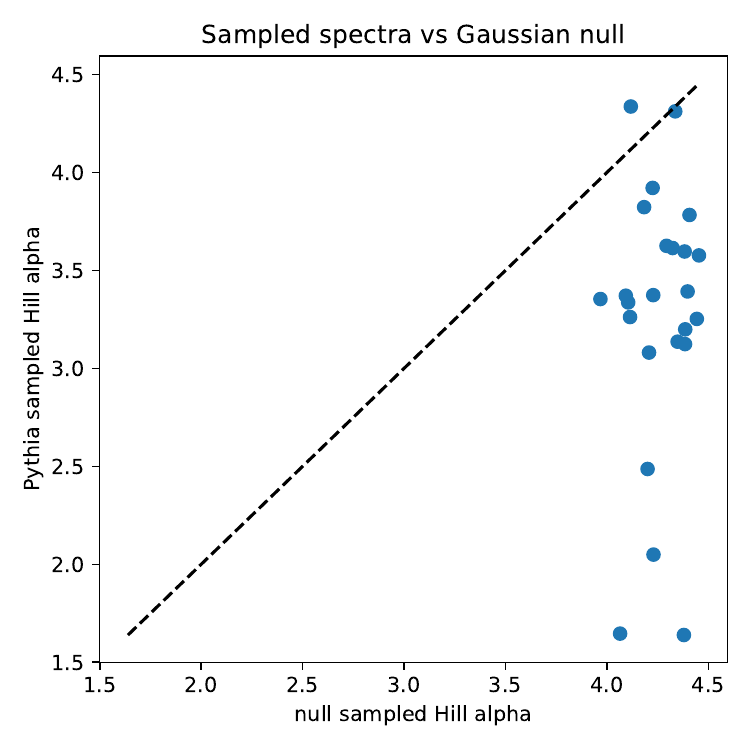}\hfill
\includegraphics[width=.32\linewidth]{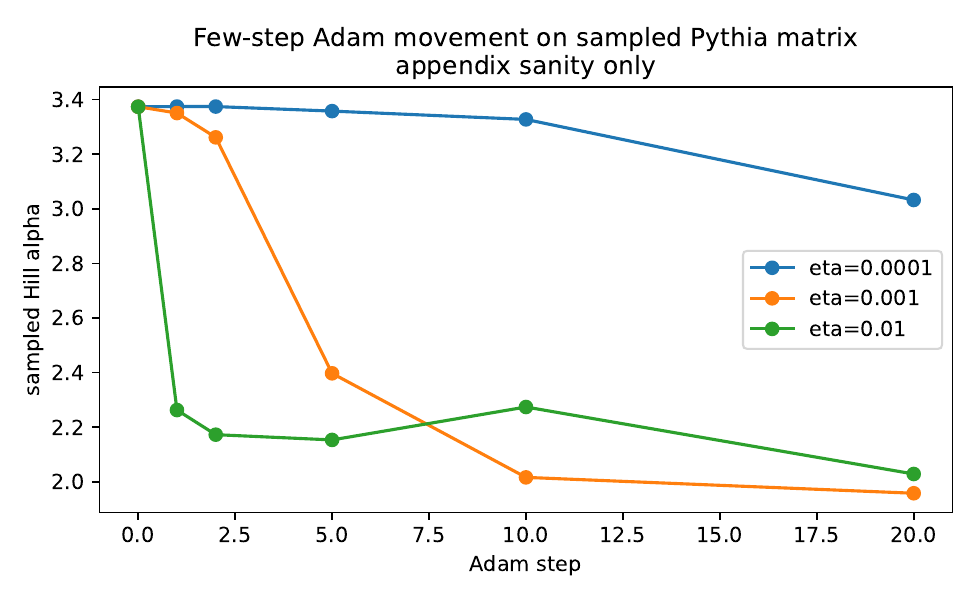}
\caption{\textbf{Pythia-70M and selected-matrix sanity checks.}  Static pretrained transformer matrices differ from matched Gaussian nulls, and the selected-matrix Adam sanity run confirms that the spectral diagnostics can be tracked on real pretrained weights.  These results do not replace the controlled onset-law experiments.}
\label{fig:pythia-sanity}
\end{figure}

\section{LLM usage statement}
Large language models were used as general-purpose assistance for drafting, editing, code organization, and consistency checking.  The authors remain responsible for all claims, proofs, experiments, numerical values, and final text.  LLMs were not used as authors and do not determine the scientific conclusions.

\section{Reproducibility statement}
All main experiments use synthetic teacher--student data generated from fixed seeds, and all reported simulation tables, figures, and audit summaries are included in the artifact package.  The appendix contains the full theorem statements and proofs, the right-censored likelihood derivation, optimizer-comparison summaries, projected-kernel robustness, onset-metric sensitivity, and real-transformer sanity checks.  The supplementary code package should include the sweep runner, configuration files, plotting scripts, and a short smoke test reproducing a small subset of the results.


\begin{thebibliography}{99}
\bibitem[Ba et~al.(2022)]{ba2022feature} Jimmy Ba, Murat A. Erdogdu, Taiji Suzuki, Zhichao Wang, Denny Wu, and Greg Yang. High-dimensional asymptotics of feature learning: How one gradient step improves the representation. In \emph{NeurIPS}, 2022.
\bibitem[Bai and Silverstein(2010)]{bai2010spectral} Zhidong Bai and Jack W. Silverstein. \emph{Spectral Analysis of Large Dimensional Random Matrices}. Springer, 2010.
\bibitem[Bietti et~al.(2022)]{bietti2022singleindex} Alberto Bietti, Joan Bruna, Clayton Sanford, and Min Jae Song. Learning single-index models with shallow neural networks. In \emph{NeurIPS}, 2022.
\bibitem[Clauset et~al.(2009)]{clauset2009power} Aaron Clauset, Cosma Rohilla Shalizi, and Mark E. J. Newman. Power-law distributions in empirical data. \emph{SIAM Review}, 51(4):661--703, 2009.
\bibitem[Cox(1972)]{cox1972regression} David R. Cox. Regression models and life-tables. \emph{Journal of the Royal Statistical Society: Series B}, 34(2):187--220, 1972.
\bibitem[Cui et~al.(2024)]{cui2024asymptotics} Hugo Cui, Luca Pesce, Yatin Dandi, Florent Krzakala, Yue M. Lu, Lenka Zdeborova, and Bruno Loureiro. Asymptotics of feature learning in two-layer networks after one gradient-step. \emph{arXiv:2402.04980}, 2024.
\bibitem[Davis and Kahan(1970)]{davis1970rotation} Chandler Davis and W. M. Kahan. The rotation of eigenvectors by a perturbation. III. \emph{SIAM Journal on Numerical Analysis}, 7(1):1--46, 1970.
\bibitem[Gurbuzbalaban et~al.(2021)]{gurbuzbalaban2021heavy} Mert Gurbuzbalaban, Umut Simsekli, and Lingjiong Zhu. The heavy-tail phenomenon in SGD. In \emph{ICML}, 2021.
\bibitem[Hill(1975)]{hill1975simple} Bruce M. Hill. A simple general approach to inference about the tail of a distribution. \emph{Annals of Statistics}, 3(5):1163--1174, 1975.
\bibitem[Hodgkinson and Mahoney(2021)]{hodgkinson2021multiplicative} Liam Hodgkinson and Michael W. Mahoney. Multiplicative noise and heavy tails in stochastic optimization. In \emph{ICML}, 2021.
\bibitem[Hodgkinson et~al.(2022)]{hodgkinson2022generalization} Liam Hodgkinson, Umut Simsekli, Rajiv Khanna, and Michael W. Mahoney. Generalization bounds using lower tail exponents in stochastic optimizers. In \emph{ICML}, 2022.
\bibitem[Kalbfleisch and Prentice(2002)]{kalbfleisch2002statistical} John D. Kalbfleisch and Ross L. Prentice. \emph{The Statistical Analysis of Failure Time Data}. Wiley, 2002.
\bibitem[Kingma and Ba(2015)]{kingma2015adam} Diederik P. Kingma and Jimmy Ba. Adam: A method for stochastic optimization. In \emph{ICLR}, 2015.
\bibitem[Kothapalli et~al.(2024)]{kothapalli2024crafting} Vignesh Kothapalli, Tianyu Pang, Shenyang Deng, Zongmin Liu, and Yaoqing Yang. Crafting heavy-tails in weight matrix spectrum without gradient noise. \emph{arXiv:2406.04657}, 2024.
\bibitem[Kothapalli et~al.(2025)]{kothapalli2025from} Vignesh Kothapalli, Tianyu Pang, Shenyang Deng, Zongmin Liu, and Yaoqing Yang. From spikes to heavy tails: Unveiling the spectral evolution of neural networks. \emph{Transactions on Machine Learning Research}, 2025.
\bibitem[Marchenko and Pastur(1967)]{marchenko1967distribution} Vladimir A. Marchenko and Leonid A. Pastur. Distribution of eigenvalues for some sets of random matrices. \emph{Mathematics of the USSR-Sbornik}, 1(4):457--483, 1967.
\bibitem[Martin and Mahoney(2021)]{martin2021implicit} Charles H. Martin and Michael W. Mahoney. Implicit self-regularization in deep neural networks: Evidence from random matrix theory and implications for learning. \emph{JMLR}, 22(165):1--73, 2021.
\bibitem[Martin et~al.(2021)]{martin2021predicting} Charles H. Martin, Tongsu Peng, and Michael W. Mahoney. Predicting trends in the quality of state-of-the-art neural networks without access to training or testing data. \emph{Nature Communications}, 12:4122, 2021.
\bibitem[Moniri et~al.(2023)]{moniri2023one} Behrad Moniri, Donghwan Lee, Hamed Hassani, and Edgar Dobriban. A theory of non-linear feature learning with one gradient step in two-layer neural networks. \emph{arXiv:2310.07891}, 2023.
\bibitem[Raj et~al.(2023)]{raj2023algorithmic} Anant Raj, Lingjiong Zhu, Mert Gurbuzbalaban, and Umut Simsekli. Algorithmic stability of heavy-tailed SGD with general loss functions. In \emph{ICML}, 2023.
\bibitem[Simsekli et~al.(2019)]{simsekli2019tail} Umut Simsekli, Levent Sagun, and Mert Gurbuzbalaban. A tail-index analysis of stochastic gradient noise in deep neural networks. In \emph{ICML}, 2019.
\bibitem[Simsekli et~al.(2020)]{simsekli2020hausdorff} Umut Simsekli, Ozan Sener, George Deligiannidis, and Murat A. Erdogdu. Hausdorff dimension, heavy tails, and generalization in neural networks. In \emph{NeurIPS}, 2020.
\bibitem[Vershynin(2018)]{vershynin2018high} Roman Vershynin. \emph{High-Dimensional Probability}. Cambridge University Press, 2018.
\bibitem[Yang et~al.(2023)]{yang2023test} Yaoqing Yang, Ryan Theisen, Liam Hodgkinson, Joseph E. Gonzalez, Kannan Ramchandran, Charles H. Martin, and Michael W. Mahoney. Test accuracy vs. generalization gap: Model selection in NLP without accessing training or testing data. In \emph{KDD}, 2023.
\bibitem[Zhou et~al.(2023)]{zhou2023temperature} Yefan Zhou, Tianyu Pang, Keqin Liu, Charles H. Martin, Michael W. Mahoney, and Yaoqing Yang. Temperature balancing, layer-wise weight analysis, and neural network training. In \emph{NeurIPS}, 2023.
\end{thebibliography}
\end{document}